\documentclass[letterpaper, 10 pt, conference]{ieeeconf}  

\IEEEoverridecommandlockouts                              

\usepackage{amsmath,amssymb,amsfonts}
\usepackage[T1]{fontenc}
\usepackage[utf8]{inputenc}
\usepackage[ruled,vlined]{algorithm2e}  
\usepackage{times}
\usepackage{adjustbox}
\usepackage{graphicx}
\usepackage{textcomp}
\usepackage{flushend}
\def\BibTeX{{\rm B\kern-.05em{\sc i\kern-.025em b}\kern-.08em
    T\kern-.1667em\lower.7ex\hbox{E}\kern-.125emX}}
\usepackage{dsfont}
\usepackage{paralist}
\usepackage[font=footnotesize]{subcaption}
\usepackage{cite}
\usepackage{tabularx}
\usepackage{cancel}
\usepackage{bm}
\usepackage{url}
\usepackage{mathtools}
\usepackage{textcomp}
\usepackage{nccmath}
\usepackage{xcolor}
\usepackage{multirow}
\usepackage{hhline}
\usepackage{booktabs}
\usepackage{makecell}
\usepackage{diagbox}
\usepackage{graphicx}
\usepackage{capt-of}
\usepackage{caption}
\usepackage{lipsum}
\usepackage{threeparttable}

\title{\LARGE \bf
Efficient Swept Volume-Based Trajectory Generation for Arbitrary-Shaped Ground Robot Navigation
}

\author{Yisheng Li*$^{1\,}$, Longji Yin*$^{1\,}$, Yixi Cai$^2$, Jianheng Liu$^1$,
Fangcheng Zhu$^1$, Mingpu Ma$^1$,\\
Siqi Liang$^1$, Haotian Li$^1$, Fu Zhang$^1$%
\thanks{\textbf{${*}$ Equal contribution.}}
\thanks{$^1$Y. Li, L. Yin, J. Liu, F. Zhu, M. Ma, H. Li and F. Zhang are with the Department of Mechanical Engineering, University of Hong Kong.}%
\thanks{$^2$Y. Cai is with Division of Robotics, Perception, and Learning, KTH Royal Institute of Technology.}
\thanks{\raggedright Email:{\tt\footnotesize \{yli385,ljyin,jianheng,zhufc,mingpu.ma,\newline liangsiqi,haotianl\}@connect.hku.hk}, {\tt\footnotesize yixica@kth.se}, {\tt\footnotesize fuzhang@hku.hk}.}
\thanks{Corresponding Author: Fu Zhang.}
}
\begin{document}

\graphicspath{{Pictures/}}  


\maketitle

\thispagestyle{empty}
\pagestyle{empty}


\begin{abstract}
    \textbf{Navigating an arbitrary-shaped ground robot safely in cluttered environments remains a challenging problem. The existing trajectory planners that account for the robot's physical geometry severely suffer from the intractable runtime. To achieve both computational efficiency and Continuous Collision Avoidance (CCA) of arbitrary-shaped ground robot planning, we proposed a novel coarse-to-fine navigation framework that significantly accelerates planning. In the first stage, a sampling-based method selectively generates distinct topological paths that guarantee a minimum inflated margin. In the second stage, a geometry-aware front-end strategy is designed to discretize these topologies into full-state robot motion sequences while concurrently partitioning the paths into SE(2) sub-problems and simpler $\mathbb{R}^2$ sub-problems for back-end optimization. In the final stage, an SVSDF-based optimizer generates trajectories tailored to these sub-problems and seamlessly splices them into a continuous final motion plan. Extensive benchmark comparisons show that the proposed method is one to several orders of magnitude faster than the cutting-edge methods in runtime while maintaining a high planning success rate and ensuring CCA.
          }
\end{abstract}

\section{INTRODUCTION}

Safe navigation of arbitrary-shaped ground robots in constrained environments is an indispensable requirement for real-world autonomy in many scenarios. For example, forklift trucks transporting long timber in the warehouse demand precise whole-body motion planning to avoid collisions in confined aisles. Traditional methods \cite{corridor-polyhedron1, corridor-sphere1,corridor-rectangle1} that conservatively approximate robots with convex shapes largely waste the navigable space, rendering them impractical for such applications. Thus, trajectory planners that explicitly model the robot’s true geometry are necessary for efficient and safe deployment in spatially constrained scenes.  


Recent frameworks like Robot-Centric Euclid Signed Distance Field (RC-ESDF) \cite{geng} and Implicit Swept Volume Signed Distance Field (SVSDF) \cite{Zhang_Wang_Xu_Gao_Gao_2023} enable precise geometry-aware robot collision assessment, overcoming the limitations of direct convex approximations. However, RC-ESDF adopts a naive A* searcher that renders a single path candidate, which ignores the possibility of other potentially feasible paths. Furthermore, its discrete sampling collision evaluation in back-end optimization risks missing the sub-resolution narrow collisions in the scene. Unlike RC-ESDF, SVSDF ensures provably accurate Continuous Collision Avoidance (CCA) \cite{SVSDF} by leveraging an iterative swept volume approach. However, the SVSDF planner also employs an A*-based searcher, which suffers the same problem as RC-ESDF's front end. Moreover, SVSDF's prohibitive runtime makes it impractical for real-world autonomous applications, which primarily stems from its monolithic back-end formulation that attempts to solve the entire whole-body robot trajectory from start to goal in a single step. In practical ground robot navigation, we observe that finding feasible navigation options rarely succeeds with a single front-end path candidate. Besides, optimizing the whole body SE(2) trajectory is necessary only in localized regions with restricted obstacles, while simpler $\mathbb{R}^2$ planning suffices elsewhere. However, both RC-ESDF and SVSDF lack a geometry-aware \textit{front end} that can provide multiple navigable options, identify critical regions, and adaptively activate whole-body optimization only where needed—leading to computational inefficiency.

\begin{figure}[tbp]
    \centering
    \includegraphics[width=0.48\textwidth]{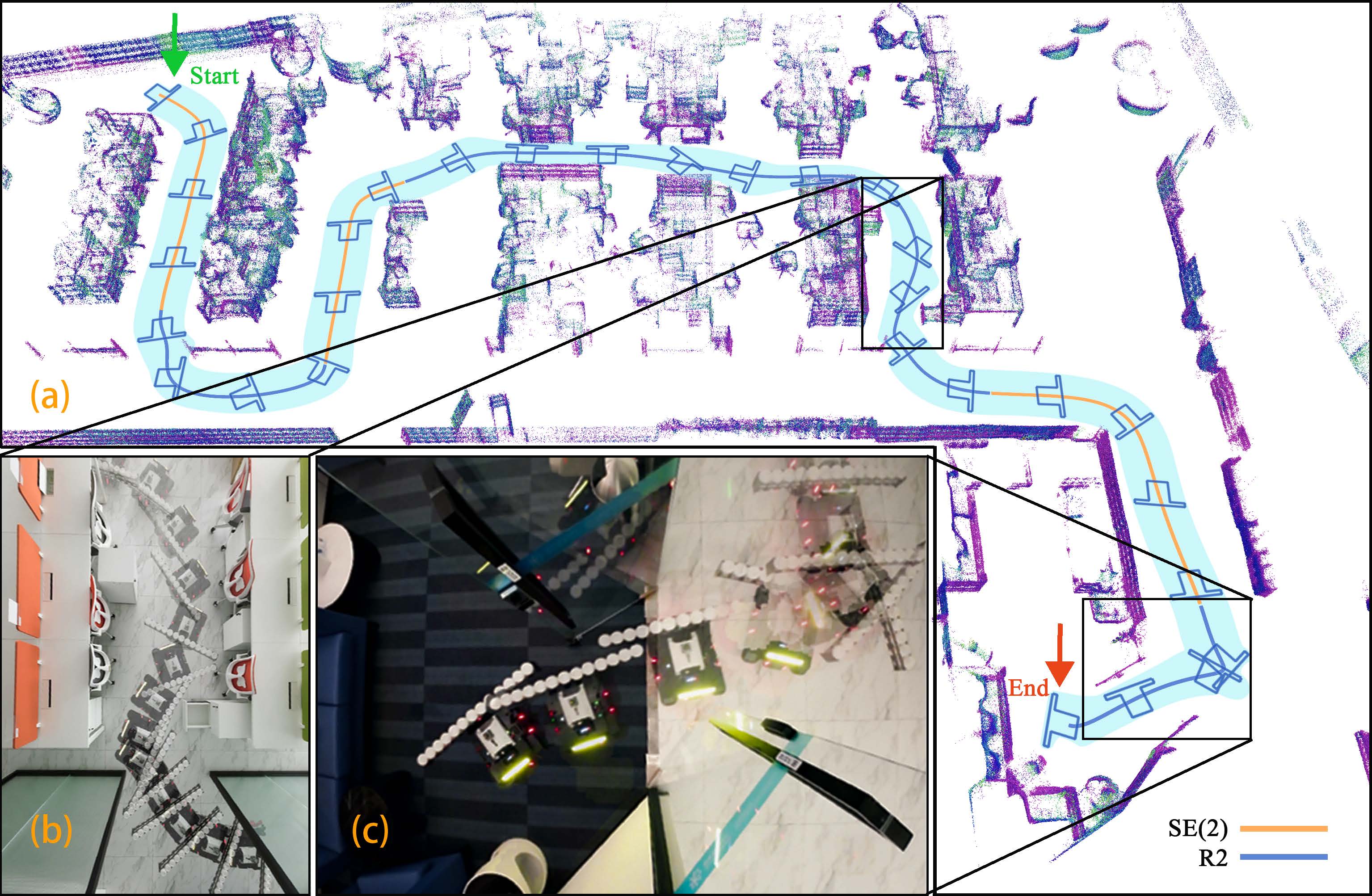}
    \captionsetup{font={footnotesize}}
    \caption{A T-Shaped delivery robot navigating a cluttered indoor environment in the real-world experiment. A whole-body trajectory is generated by the proposed framework to ensure precise continuous collision avoidance.}
    \label{fig:coverpage}
    \vspace{-7.8pt}
\end{figure}

To address the problems above, this paper proposes a coarse-to-fine navigation framework for arbitrary-shaped ground robots, enabling optimal navigation under continuous collision avoidance (CCA) in real-world complex scenarios with only minor computational overhead. Firstly, we answer how to generate multiple navigation options in an obstacle-dense environment. We propose a topological path discovery approach with minimum safety margins and a shape-tight-coupling path refinement strategy, enabling the discovery of narrow yet potentially feasible paths that single-path methods would prematurely eliminate. Secondly, based on these topological path candidates, we answer the question of how to identify the critical regions (\textit{i.e.}, high-risk regions) to accelerate the trajectory generation. We propose an SE(2) motion sequence generation that employs fast collision check to accurately decouple the optimization problem, locating high-risk regions such as narrow passages as SE(2) problems (requiring precisely swept volume-aware) while processing low-risk regions such as open spaces as simpler $\mathbb{R}^2$ problems. Finally, a back-end optimization paradigm is designed to efficiently generate CCA-guaranteed trajectories according to the problem classification, thereby increasing the overall computational efficiency while maintaining safety. Experiments demonstrate excellent performance with both T-shaped and L-shaped robots in simulated environments as well as with a real-world T-shaped drink delivery robot, illustrated in Fig.~\ref{fig:coverpage}. To summarize, the contributions of this paper are as follows:
    
    1) We present a topology generation module that novelly enables geometry-aware shortening on detoured topological paths, rendering high-quality topology candidates.
    
    2) We propose a discrete SE(2) motion sequence generation that enables geometry-aware local path segment adjustment, partitioning the paths into SE(2) sub-problems and efficient $\mathbb{R}^2$ sub-problems.

    3) We design a back-end trajectory optimization paradigm that generates the global Continuous Collision Avoidance (CCA) trajectory for arbitrarily shaped robots, ensuring low time consumption.

    4) Extensive benchmark comparisons and real-world experiments validate the performance of our proposed method, demonstrating its computational efficiency and high success CCA rate compared to state-of-the-art baselines. 

    5) We will open-source our algorithm to the community to support further research and practical applications.\footnote{\url{https://github.com/hku-mars/ESV-Planner}}

\begin{figure*}[!h]
    \vspace{+3pt}
    \centering
    \includegraphics[page=1, draft=false, width=0.77\textwidth]{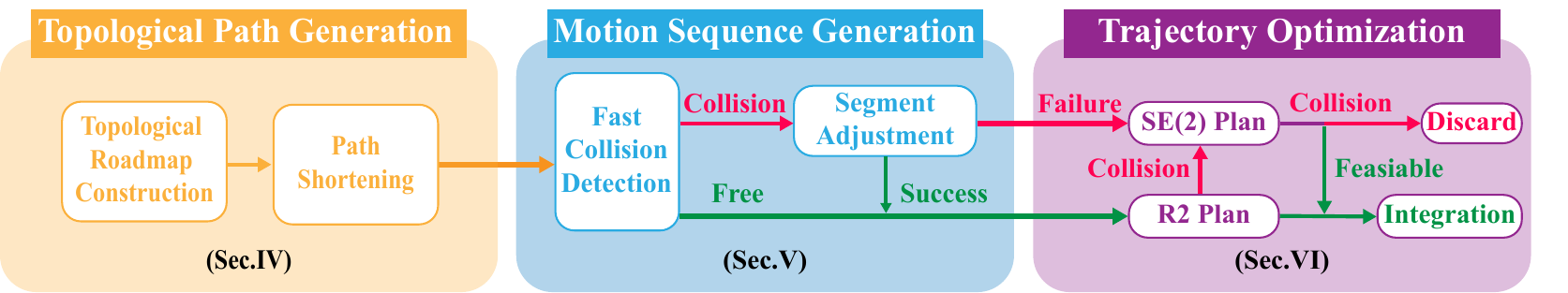}
    \captionsetup{font={footnotesize}}
    \caption{The overview of our proposed planning framework.}
    \label{fig:framework}
    \vspace{-0.2cm}
\end{figure*}

\section{RELATED WORKS}

In robot motion planning, existing works \cite{bubble-planner}, \cite{kong_drone_sphere} usually model the robot's geometry as a mass point and inflate the obstacle based on its radius to construct configuration space for collision avoidance. Liu \textit{et al.} \cite{Liu_Mohta_Atanasov_Kumar_2018} encodes the robot shape as an ellipsoid to search the safe path for drones. Euclidean Signed Distance Fields (ESDF) and corridor-based methods are widely employed to construct collision-aware cost functions for whole-body motion planning trajectory optimization. 

The ESDF serves as a map representation that encodes the distance to the nearest obstacle surface. In this approach, the robot’s geometry is often approximated as a sphere \cite{ego_planner}, \cite{fast_planner} to simplify distance queries. Collision-free motion is guaranteed by ensuring the ESDF value (\textit{i.e.}, the minimum distance to obstacles) at each trajectory sample point exceeds the radius of the corresponding sphere. While computationally efficient, this spherical approximation may lead to overly conservative trajectories, as the spheres only coarsely enclose the robot’s true geometry. In contrast,  corridor-based methods construct trajectories by defining collision-free regions as convex geometric primitives, including polyhedra \cite{corridor-polyhedron1}, spheres \cite{corridor-sphere1}, and axis-aligned bounding boxes \cite{corridor-rectangle1}. Li \textit{et al.}  \cite{dual-sphere-parking} employs multiple spheres to approximate the robot's shape, enabling a more flexible and accurate representation of its geometry within the collision-free corridors. These primitives impose linear constraints on the trajectory optimization problem, ensuring the robot’s motion remains within the corridor. To account for the robot’s spatial footprint, the corridor framework often models the robot as an ellipsoid \cite{yunfan-wholebody}, a union of spheres \cite{dual-sphere-parking}, \cite{Li_Ouyang_Zhang_Acarman_Kong_Shao_2021} and enforces its containment within the corridor’s convex polyhedra. All the above-mentioned methods share a core principle: generating a convex polytope that encapsulates the robot’s geometry to guarantee collision avoidance. However, ensuring continuous collision-free trajectories while preserving the accurate representation of a robot's non-convex geometric shape remains a significant challenge.

Inspired by the flexibility of the ESDF in representing arbitrary obstacle geometries, Geng \textit{et al.} \cite{geng} proposed a novel approach that inverts the traditional ESDF paradigm: instead of computing distances from obstacles, the method evaluates distances from the robot’s surface to obstacles. By reformulating the ESDF around the robot’s geometry, the framework actively ``pushes'' the robot away from obstacles during trajectory optimization. However, this method faces two critical limitations. First, like traditional ESDF, the robot-centric ESDF relies on discrete sampling of the robot’s surface along trajectory which may results in ``tunnel effect''. Second, the gradient direction derived from the robot-centric ESDF does not reflect the optimal escape direction for the robot.


To address these issues, SVSDF\cite{SVSDF} innovatively integrates the concept of swept volume into trajectory optimization. It represents the robot’s surface as the zero-level set of an implicit SDF and integrates swept volume to model the space-time continuum of its motion to achieve resolution-independent CCA. Additionally, the SVSDF is calculated by finding the radius of the smallest sphere at an obstacle point that is tangent to the swept volume boundary which is always along the optimal gradient for collision avoidaance. However, precisely considering the robot's actual geometry in a single optimization problem from start to end for the continuous collision-free trajectory is time-consuming. It causes much pressure for the optimizer to converge to a satisfied minimum. Thus, we present an adaptive coarse-to-fine trajectory generation paradigm, accurately representing the robot's geometry only when it is needed, and using simpler planners elsewhere to speed up the process.


\section{Framework Overview}

Our proposed framework, as shown in Fig.~\ref{fig:framework}, consists of three stages: topological path generation, SE(2) motion sequence generation, and trajectory generation. \textbf{1)} The topological path generation creates multiple topologically distinct paths and shortens them by a geometry-aware shortcut algorithm in a map inflated by the inscribed circle radius of the robot's geometry, which fully explores the potential paths to achieve high reversibility (Sec.\ref{sec:sectionA}). \textbf{2)} SE(2) motion sequences are generated from the topological paths, which enable geometry-aware collision path adjustment and find a sequence of full-state (position and orientation) collision-free robot configuration while determining the optimization mode (\textit{i.e.}, SE(2) or $\mathbb{R}^2$) for the subsequent back-end trajectory optimization (Sec.\ref{sec:sectionB}). \textbf{3)} Trajectory generation optimizes the SE(2) and $\mathbb{R}^2$ subproblems separately. Any SE(2) collision results are then discarded, while $\mathbb{R}^2$ collisions are re-optimized by the SE(2) planner if they occur. Finally, these optimized trajectory segments are combined into a continuous final navigable option Sec.\ref{sec:sectionC}.

\section{Topological Path Generation} \label{sec:sectionA}
\begin{figure}[hbp]
    \vspace{+2pt}
    \centering
    \includegraphics[width=0.48\textwidth]{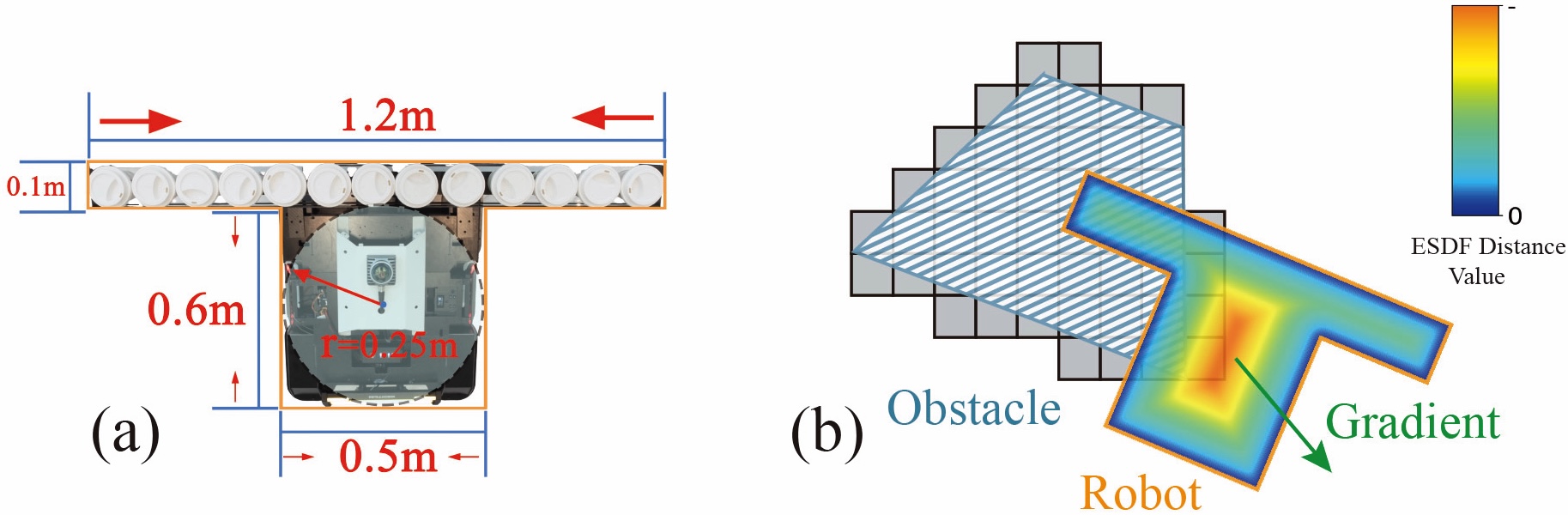}
    \captionsetup{font={footnotesize}}
    \caption{(a) Actual experiment robot with dimension marked, where the inscribed circumference of the robot is shown. (b) The robot body frame ESDF is built inside the robot geometry. The sum of the gradient (green arrow) can be used to avoid the obstacle.}
    \label{fig:roboesdf}
    \vspace{-10pt}
\end{figure}

To maximize the exploration of various topological paths in a narrow space while maintaining efficiency, in this stage, we use the minimum inflation margin which is the inscribed circumference of the robot as shown in Fig.~\ref{fig:roboesdf}(a), to inflate the occupancy grid map, representing the coarse navigability at this initial stage. After that, we construct topological roadmaps based on \cite{topology} with a rough consideration of the robot's shape.

\begin{algorithm}
    \caption{Path Shortcut}
    \LinesNumbered
    \KwIn{Original path $P_i$, robot geometry $\mathcal{M}$ }
    \KwOut{Simplified SE(2) path $P_s$}
    $P_d \leftarrow \textbf{UniformDiscretize}(P_i); P_s \leftarrow \{p_1\}$\;
    \ForEach{$p_d \in \mathcal{P}_d$}{
        \If{$\neg \textbf{IsVisible}(P_s.back(), p_d)$}{
            $p_{c} \leftarrow \textbf{CollisionPoint}()$\;
            $\mathcal{M}_{yaw} \leftarrow \textbf{GetRoboState}(p_{c},\mathcal{M})$\;
            $p_{new} \leftarrow \neg \textbf{Safe}$\;
            \While{$\neg \textbf{Safe}(p_{new})$ $\land$ $N_{attempt}$ $<$ $N_{max}$}{
            $X_{obs} \leftarrow \textbf{ExtractObstacles}(\mathcal{M}_{yaw})$\;
            $\mathcal{SDF}^{\mathcal{M}} \leftarrow \textbf{ESDF}(\mathcal{M}_{yaw},X_{obs})$\;
            $\theta_{new},p_{new} \leftarrow \textbf{PushAway}(\mathcal{M},\mathcal{SDF}^{\mathcal{M}})$\;
                        $\mathcal{M}_{yaw} \leftarrow \textbf{UpdateState}(\mathcal{\theta}_{new},p_{new},\mathcal{M})$\;
            }
            $P_s.\textbf{emplace}(\theta_{new},p_{new})$\;
        }
    }
    $P_s.\textbf{emplace}(P_d.back())$\;
    \Return $P_s$
\end{algorithm}

\begin{figure}[h]
    \centering
    \includegraphics[width=0.48\textwidth]{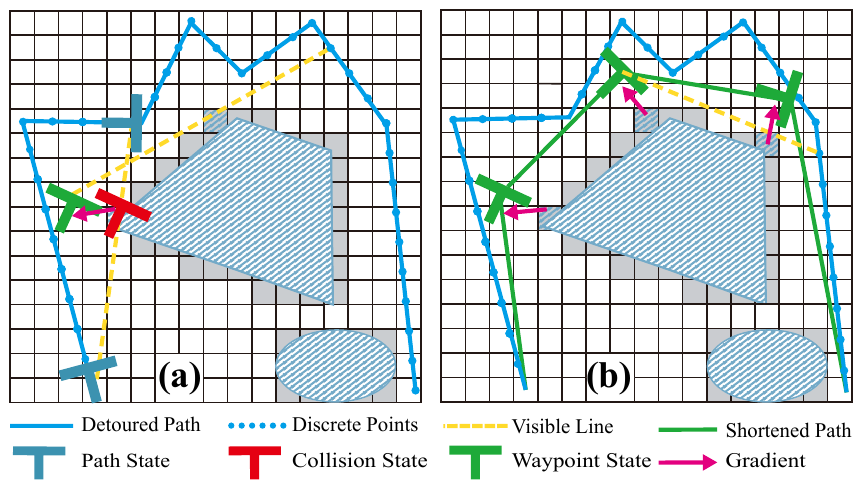}
    \captionsetup{font={footnotesize}}
    \caption{A detoured and long path is shortened based on the robot's actual geometry. (a) At each discretized point, the obstacle that blocked its visibility to the last point is pushed to generate a new waypoint and stored with a safe orientation (green T-shaped). (b) The final simplified topological path (green line). }
    \label{fig:PathShortcut}
    \vspace{-10pt}
\end{figure}

In \cite{topology}, redundant zigzag paths in obstacle-dense regions are shortened while treating the robot as a mass point and ignoring the actual shape. This will result in missing regions that can be passed in a particular orientation. Therefore, we redesigned the shortcut process to generate a refined topologically equivalent shortcut path $P_s$ considering the any-shape robot's physical geometry (refer to Alg. 1 and Fig.~\ref{fig:PathShortcut}). Initially, we denote the original detoured path (blue line) found by the depth-first search from the roadmap as $P_i$ and the expected final simplified path (green line) as $P_s$. Then, the algorithm uniformly discretizes $P_i$ into a set of points $P_d = \{p_1, p_2, ..., p_n\}$ (blue point in the path) based on the grid map resolution and initializes the simplified path $P_s$ with the start point $p_1$ (Line 1). For each point $p_d$ in $P_d$, the algorithm performs visibility checking between $P_s.back()$ and $p_d$  as in \cite{topology} (Line 2). When visibility is blocked by an obstacle, the algorithm identifies the obstruction point as $p_{c}$ (Line 4).

Then, we expect to push the obstructed configuration in $p_c$ away from the occupancy to generate new collision-free waypoints. Inspired by \cite{geng}, to handle the arbitrary-shaped robot, we construct body-frame ESDF inside the robot geometry $\mathcal{M}$ centered at obstruction point $p_c$ to render the obstacle-avoiding direction, as shown in Fig.~\ref{fig:roboesdf}(b). The robot’s initial orientation at the obstruction point $p_c$ (red T-shape) is important to construct the appropriate body-frame ESDF for guiding collision point adjustment. It is determined by applying Cubic Spline Interpolation between the initial and final orientation at $P_s.back()$ and $p_{d}$ (blue T-shape), respectively, and then storing the resulting orientation with the geometric representation and position in $\mathcal{M}_{yaw}$ for generating the ESDF (Line 5). we build an Axis-Aligned Bounding Box (AABB) to extract all nearby obstacle points $X_{obs}$ relative to $p_c$ (Line 8).
The SDF value at any position $x_{obs}$ in $X_{obs}$ is defined as the minimum distance from it to the nearest robot surface, with positive values indicating the robot's exterior and negative values indicating its interior, which can be efficiently computed via a generalized winding number method \cite{generalizedwindingnumber} within LIBIGL\footnote{\url{https://libigl.github.io/}} (Line 9). The gradient in the robot local frame $\nabla\mathcal{SDF}_{robot}^{\mathcal{M}}$ is defined as the direction of the steepest ascent to the nearest robot surface from the query point $x_{obs}$ and computed using central finite differences of distance values between adjacent grid cells. Subsequently, the gradient of the robot SDF in the world frame $\nabla\mathcal{SDF}_{world}^{\mathcal{M}}$ which directs away from the obstacle is obtained with a proper coordinate transform as follows:
\begin{equation}
    \nabla\mathcal{SDF}_{world}^{\mathcal{M}} = {R}^{-1}(p_c) \cdot \nabla\mathcal{SDF}_{robot}^{\mathcal{M}},
\end{equation}
where ${R}^{-1}(p_c)$ is the inverse of the robot's rotation matrix at $p_c$.  The norm and direction of the gradient (pink arrow) can push the center of robot ESDF in obstruction point $p_c$ away from the obstacle with a magnitude based on the SDF value and minimum safe margin (Line 10). The resultant position after applying SDF-based steering once might remain collided with the confined environments. Thus, we repeat this process within the maximum allowed attempts (Line 7). If the robot collides after reaching the limitation attempts, the last $p_{new}$ along with its orientation $\theta_{new}$ is temporally designated as a new waypoint for further SE(2) refinement (Line 12). The refining process continues until the last point is reached. After that, uniform visibility deformation (UVD) \cite{topology} is implemented to distinguish topological equivalence and eliminate redundant paths effectively. Finally, Alg. 1 generates a new shortened and smoother path consisting of a series of waypoints with feasible orientation stored according to the actual geometry of the robot while preserving the topology attribute.
\section{SE(2) Motion Sequence Generation} \label{sec:sectionB}
After the first stage in Sec.~\ref{sec:sectionA}, a series of refined topology paths has been rendered in the occupancy map inflated by the robot's inscribed circumference. The topology paths are already feasible for planners with the conventional mass point model. However, for geometrically complex robots, the paths still risk environmental collisions during navigation, as the path segments between the adjacent waypoints lack geometry-aware collision avoidance. 

\begin{figure}[htbp]
    \centering
    \includegraphics[page=1, draft=false, width=0.45\textwidth]{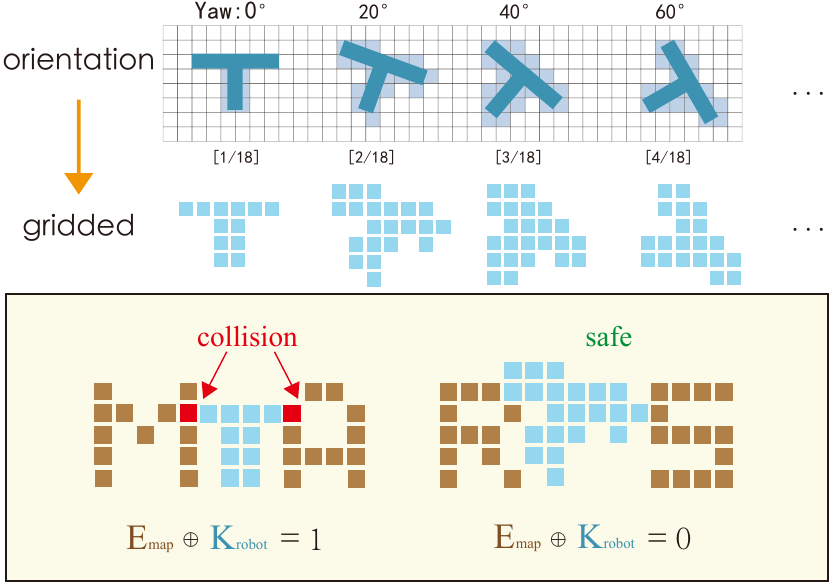}
    \captionsetup{font={footnotesize}}
    \caption{A T-shaped robot detects a collision with the environment, where the robot kernel discretizes 18 gridded configurations at 20-degree clockwise rotational increments.}
    \label{fig:robotKernel}
\end{figure}

At this stage, we generate a sequence of full-state collision-free discrete robot configurations (called an SE(2) motion sequence) for each topology path, which stores the robot's position, the orientation of the robot geometry template, and the back-end optimization type (high-risk or low-risk). These SE(2) motion sequences act as high-quality initial values for downstream back-end trajectory generation, directly affecting the computational efficiency of the optimization process. Specifically, the orientation of the robot geometry in the SE(2) motion sequence is stored in a grid representation within the prescribed resolution to accelerate collision detection, as shown in Figure~\ref{fig:robotKernel}. We name the grid representations as ``robot kernel'' in the following descriptions, denoted as $\mathcal{K}_{robot}$. Besides, the back-end optimization type is determined within the sequence generation process, in which the path is partitioned and labeled as either high-risk zones that demand SVSDF-based SE(2) optimization or low-risk regions where simpler $\mathbb{R}^2$ optimization suffices. 

We introduce the motion sequence generation in Alg.2. The algorithm begins by generating a robot kernel $\mathcal{K}_{robot}$ from the robot's geometry (Line 1). Subsequently, we review the path segments $S$, which are the connections between each pair of two neighboring waypoints in the topological paths from Alg. 1. Each path segment is uniformly discretized into a point set $P_d$ according to grid map resolution (Line 3) and each point $p$ in $P_d$ is reviewed as follows.

\begin{figure}[!t]
    \centering
    \includegraphics[width=0.48\textwidth]{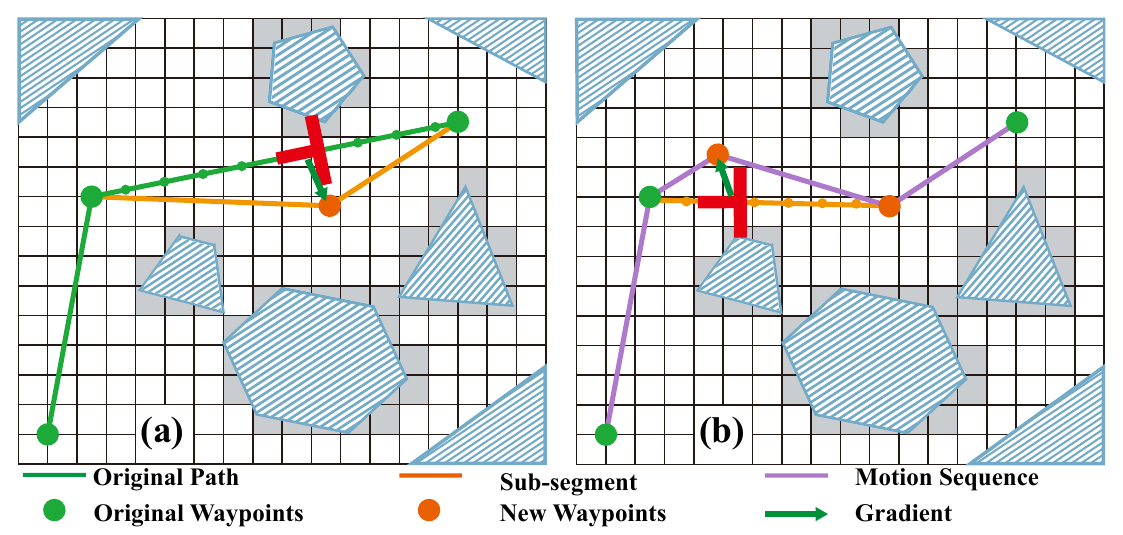}
    \captionsetup{font={footnotesize}}
    \caption{Two phase of a collision-prone SE(2) motion sequence is refined. a) A collision occurs on the original path (red T-shape) at a point $p$. The point is then steered to a free position (orange point) which is then connected to existing ones (large green points).  b) Child segments with remaining collisions undergo repeated adjustment until achieving a fully collision-free SE(2) motion sequence (purple line).}
    \label{fig:PathRefine}
    \vspace{-8pt}
\end{figure}

We first find a collision-free orientation at $p$ using the function \textit{SafeYaw()}, which performs a Boolean convolution operation conducted between the occupancy map $E_{map}$ and the robot kernel $\mathcal{K}_{robot}$, as shown in Figure~\ref{fig:robotKernel}. This process is computationally efficient since the robot geometry template is pre-loaded in memory. If this robot template with initial orientation is detected colliding, \textit{SafeYaw()} sequentially explores other collision-free templates $\mathcal{R}_{free}$ within a limited angular range (Line 6). Once  \textit{SafeYaw()} returns a feasible template, the full-state configuration is marked as low-risk and is temporarily stored in the motion sequence buffer $E_{tmp}$ (Line 18). Otherwise, when no available template is found at point $p$ (Line 7), there may still be open space on the other side of the obstacle that the robot can easily pass through without considering full-state SE(2) planning. This is the rationale of conducting \textit{SegAdjust($S,p$)}, as illustrated in Fig~\ref{fig:PathRefine} and explained below.

 
When \textit{SafeYaw()} fails at specific $p$ due to obstacle collision, the algorithm will call a \textit{SegAdjust()} function to examine if the segment can be locally adjusted to have a feasible template (Line 8). Firstly, we steer the original infeasible $p$ away from the obstacle using the body-frame ESDF (\textit{PushAway()} in Alg.1) and denote the new point as $p'$ (orange dot in Fig.~\ref{fig:PathRefine}). Afterward, we connect the start and the end of the original segment $S$ with $p'$ to form two new sub-segments (shown in Fig.~\ref{fig:PathRefine}(a)). Then, on each sub-segment, we repeat the discrete collision detection, ESDF-based steering, and sub-segment splitting process until the new segment series connecting the start and the end of $S$ are all feasible. In that case, these new segments are adopted to replace the original $S$, and all robot configurations along the new segments are emplaced and marked as low-risk (Line 10-13, shown in Fig.~\ref{fig:PathRefine}(b)). Conversely, if one of the sub-segments cannot survive the collision, the whole \textit{SegAdjust()} is discarded, and point $p$ on original $S$ is marked as high-risk (Line 16). Finally, the motion sequence buffer $E_{tmp}$ will be added to the final motion sequence $E_{final}$ (Line 19). 

\begin{algorithm}
    \caption{SE(2) Motion Sequence Generation}
    \LinesNumbered
    \KwIn{Original $\mathbb{R}^2$ topo path $P_r = {w_1, w_2, ..., w_i}$, robot geometry $\mathcal{M}$ }
    \KwOut{$\mathbb{SE}(2)$ Sequence $E_{final}$}
    $\mathcal{K}_{robot} \leftarrow \textbf{GenerateRobotKernel}(\mathcal{M})$\;
    $E_{tmp} \leftarrow \emptyset$ \;
    \ForEach{$S \in \textbf{GetPathSegments}(P_r)$}{
    $P_{d} \leftarrow \textbf{UniformDiscretize}(S)$\;
    \ForEach{$p \in P_{d}$}{
    $\mathcal{R}_{free} \leftarrow \textbf{SafeYaw}(p,\mathcal{K}_{robot})$\;
    \If{$\mathcal{R}_{free} = \emptyset$}{
    $S_{adj} \leftarrow \textbf{SegAdjust}(S,p)$\;
    \If{$S_{adj} \neq \emptyset $}{
    $P_{adj} \leftarrow 
    \textbf{UniformDiscretize}(S_{adj})$ \;
    $E_{tmp} \leftarrow \emptyset$ \;
    \ForEach{$p_{adj} \in P_{adj}$}{
    $E_{tmp}.\textbf{emplace}(p_{adj},\mathcal{R}_{free}, \textbf{LowRisk})$
    }
    \textbf{break}\;
    }
    \Else{
    $E_{tmp}.\textbf{emplace}(p,\mathcal{R}_{free}, \textbf{HighRisk})$\;
    }
    }
    \Else{
    $E_{tmp}.\textbf{emplace}(p,\mathcal{R}_{free}, \textbf{LowRisk})$\;
    }
    }
    $E_{final} \leftarrow E_{final} \cup E_{tmp}$ \;
    $E_{tmp} \leftarrow \emptyset$ \;
    }
    \Return $E_{final}$
    \end{algorithm}

In summary, Alg.2 generates an SE(2) motion sequence for each topological path candidate that stores the robot position, orientation, and the corresponding optimization option.
Meanwhile, narrow environmental regions have been efficiently classified into high-risk and low-risk regions while decoupling the motion planning problem from the full SE(2) space to simplified $\mathbb{R}^2$ optimization. This reduces computational complexity while preserving trajectory integrity. In the following section, the back end will optimize the sequence to continuous and smooth SE(2) and $\mathbb{R}^2$ trajectories accordingly.
\section{Trajectory Optimization} \label{sec:sectionC}
\subsection{Trajectory Generation Strategy}

Although Alg.2 localizes high-risk areas in the environment by generating SE(2) motion sequences for each original topological path candidate, the geometric representation in the robot kernel is not sufficiently accurate because it is gridded. This leads to the fact that collisions and obstacles detected using the robot kernel do not necessarily prove that a non-convex geometry robot is physically infeasible. Conversely, path segments examined by discretization do not detect sub-resolution collisions and ensure continuous collision-free collisions. Therefore, in this section, we employed the SVSDF planner as our SE(2) sub-problem solver to enable precise geometry-aware robot collision assessment.

\begin{figure}[htbp]
    \vspace{+2pt}
    \centering
    \includegraphics[width=0.45\textwidth]{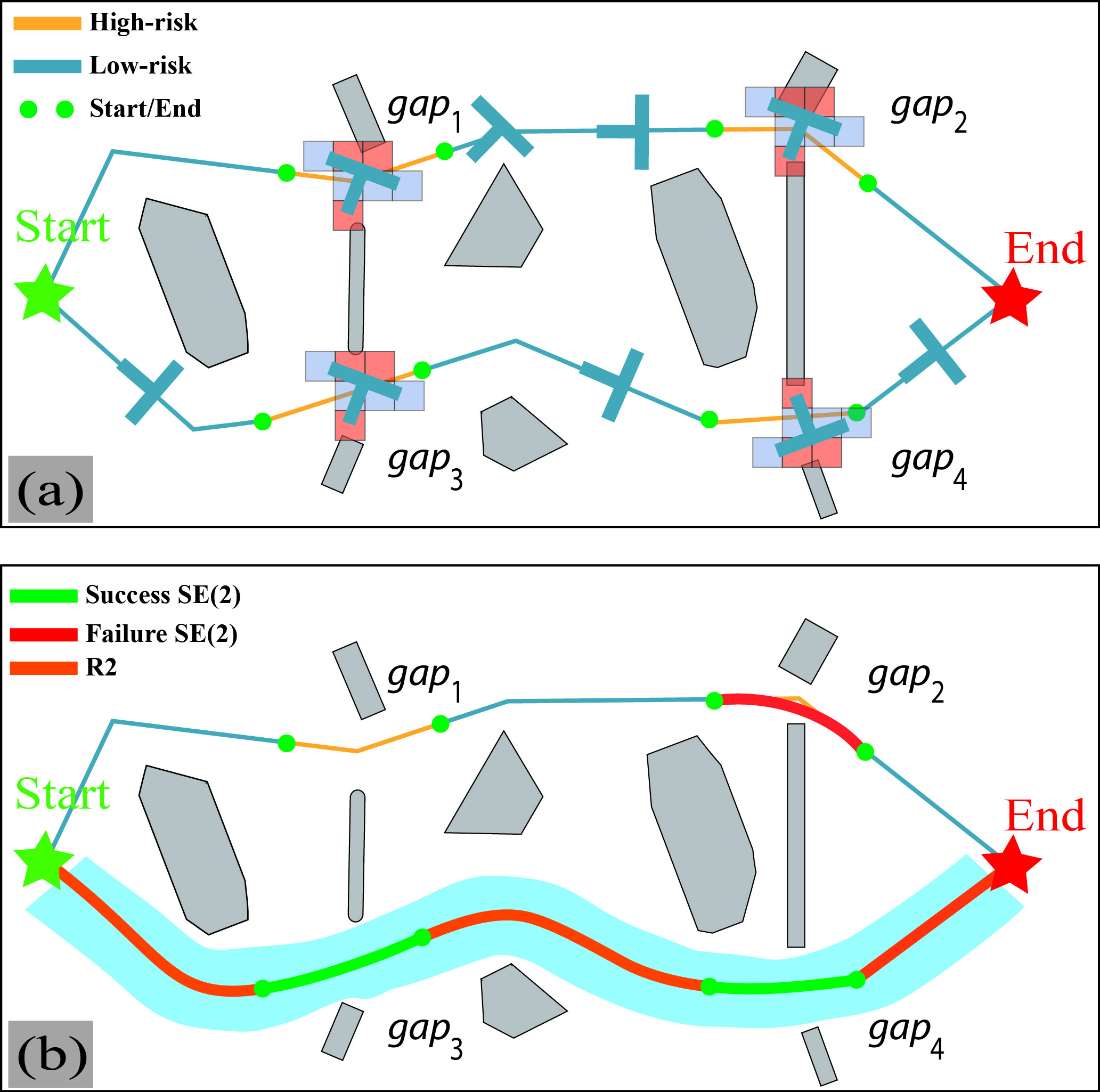}
    \captionsetup{font={footnotesize}}
    \caption{Two refined SE(2) motion sequences for a T-shaped robot navigating an obstacle-dense zone are visualized. (a) Robot kernel collides with obstacles in all \textbf{$gaps$}. (b) A complete, collision-free trajectory composed of two successfully optimized SE(2) segments connected by three $\mathbb{R}^2$ segments (orange curve). The successfully SE(2) optimized trajectory pieces in \textbf{$gap_3$} and \textbf{$gap_4$} are shown in green color, while failed in \textbf{$gap_2$} indicates with red.}
    \vspace{-4pt}
    \label{fig:TrajOp}
\end{figure}

Our method prioritizes verifying traversability through high-risk regions before optimizing the remaining trajectory segments in $\mathbb{R}^2$ sub-problem. If all the SE(2) sub-problems in the same trajectory result in safety, the left open space region is subsequently optimized by the $\mathbb{R}^2$ planner. After both SE(2) and $\mathbb{R}^2$ successfully generate and splice to a complete navigable trajectory, we perform a final precise collision evaluation on the entire trajectory where any unsafe trajectory piece from $\mathbb{R}^2$ results is re-optimized by SE(2) planner. If one of the SE(2) is optimized but remains in a collision, the left SE(2) segments are discarded and the algorithm continues to evaluate alternative candidate topological paths. Finally, the minimum control effort (most smoothness) trajectory is selected as the final navigable option for the ground robot.

We take $gap_1$-$gap_4$ in Fig.~\ref{fig:TrajOp} as an example to better explain our method. Initially, segment extraction isolates high-risk robot configuration (orange path in Fig.~\ref{fig:TrajOp} (a)) from the complete SE(2) motion sequence. This extraction generates a series of $N$ candidate SE(2) sub-problems in $gap_1$-$gap_4$, each comprising a motion sequence of full-state robot configurations within a high-risk region. We then start to optimize each extracted SE(2) segment sequentially by SVSDF planner, referring to Eq.~\ref{eq:SE2} in Section~\ref{sec:TrajFormulation}. For the top path in Fig.~\ref{fig:TrajOp}(a), the algorithm cannot find a collision-free trajectory (red curve) in $gap_2$ after SE(2) is optimized. This results in neglecting the trajectory generation in $gap_1$, and this path candidate is discarded, as shown in Fig.~\ref{fig:TrajOp}(b). Conversely, in the case of SE(2) sub-problems optimized successfully (green curve) in both $gap_3$ and $gap_4$, the swept volume (blue pipeline) is performed continuously by a precise collision check. Then, this trajectory is selected as the final navigable option.

\subsection{Trajectory Formulation} \label{sec:TrajFormulation}
In this work, we adopt MINCO \cite{MINCO}, a piece-wise polynomials representation of the trajectory:


\begin{equation}
\mathbf p(t)\!=\!
\begin{cases}
\mathbf p_1(t)= \mathbf c_{1}^T\beta\left(t-0\right) & 0 \leq t<T_1 \\
\vdots & \vdots \\
\mathbf p_M(t)\!=\! \mathbf c_{M}^T\beta\left(t-T_{\!M\!-\!1}\right) \!\!\!& T_{\!M\!-\!1} \!\leq\! t\!<\!T_{\!M}
\end{cases}
\end{equation}
Here, the trajectory \(p(t)\) is an \(m\)-dimensional polynomial composed of \(M\) segments, each with a degree \(N = 2s - 1\), where \(s\) represents the order of the corresponding integrator chain. \(c_i \in \mathbb{R}^{(N + 1) \times m}\) denotes the coefficient matrix of ${M}$ pieces, \(\beta(t) = [1, t, \ldots, t^N]^T\) represents the natural basis,and \(T_i = t_i - t_{i-1}\) is the time duration of the \(i\)-th segment. The robot's state can be easily obtained by taking the derivative of the polynomial trajectory. In this paper, we use the degree \(N = 5\) to ensure the continuity up to snap in adjacent pieces.

As previously outlined, the SE(2) sub-problem solver must holistically optimize trajectory quality by minimizing control effort (smoothness), and rigorously enforcing continuous safety constraints. To achieve this balance, we formulate the following cost function for the SE(2) trajectory optimization:
\begin{equation} \label{eq:SE2}
    \min_{c, T} \text{Cost}_{\text{SE(2)}} = \lambda_m J_m + \lambda_t J_t + \lambda_s J_s+ \lambda_d J_d.
\end{equation}
where the terms \( J_m \), \( J_t \), \( J_s \), and \( J_d \) are the smoothness, total time, safety, and dynamics penalties, respectively.
 \( \lambda_m \), \( \lambda_t \), \( \lambda_p \), and \( \lambda_R \) are their corresponding weights. 

Considering the computationally intensive nature of querying the SVSDF on the entire trajectory, we assume sufficient inherent safety in $\mathbb{R}^2$ segments to disregard safety constraints in low-risk $\mathbb{R}^2$ sub-problems. This assumption is supported by our carefully designed preprocessing step where Alg.1 refines the geometric-aware topological path and Alg. 2 performs redundant discrete collision checks. The $\mathbb{R}^2$ trajectory optimization then focuses on:
\begin{equation}\label{eq:R2}
\min_{c, T} \text{Cost}_{\text{$\mathbb{R}^2$}} = \lambda_m J_m + \lambda_t J_t + \lambda_p G_p + \lambda_R G_R.
\end{equation}
Here, the terms \( J_m \), \( J_t \), \( G_p \), and \( G_R \) represent the smoothness, total time, position, and pose residual penalties, respectively. The position residual \( G_p(t) \) is defined as follows, utilizing the \( C^2 \)-smoothing function \( L_\mu [\cdot] \):
\begin{equation}
G_p(t) = L_\mu \left[ \| p(t) - p^{i(t)} \|^2 \right],
\end{equation}
where \( \| \cdot \|^2 \) denotes the square of the Euclidean norm of a vector. The function \( i(t) \) maps to the index of the discretized key node based on the SE(2) motion sequence. The rotation residual \( G_R(t) \) is defined as:

\begin{equation}
G_R(t) = L_\mu \left[ \| R(t)^{-1} R^{i(t)} - I \|^2_F \right].
\end{equation}

Here, \( \| A \|^2_F \) represents the Frobenius norm of matrix \( A \), which can be expressed as \( \text{tr} (A^T A) \) using the matrix trace. Essentially, \( G_R(t) \) quantifies rotation similarity residuals, and \( G_p(t) \) evaluates position similarity residuals. These optimization components are designed to ensure the closeness between optimization result trajectory and motion sequences.

\section{EXPERIMENTS}
\subsection{Benchmark Comparison}
\subsubsection{Baselines} We evaluate the proposed method against two state-of-the-art any-shape robot motion planners: SVSDF \cite{SVSDF} and RC-ESDF \cite{geng}. Both baselines deploy A* as their front-end global path search but differ fundamentally in collision-free orientation storing. SVSDF extends A* to explore robot kernel position and orientation during node expansions, storing the full SE(2) state at each node. In contrast, RC-ESDF adopts traditional A* without considering orientation in this stage, parameterizing trajectories as uniform B-splines. Neither baseline inherently supports topological diversity paths. For fairness, we unify all methods using MINCO \cite{MINCO} for trajectory representation and generate reference topological paths via Zhou’s method \cite{topology}. RC-ESDF is guided by $\mathbb{R}^2$ paths with collision avoidance. At the same time, SVSDF initializes SE(2) optimization by discretizing topological paths and solving for feasible orientations at each point.
\begin{figure}[tbp]
    \centering
    \includegraphics[width=0.45\textwidth]{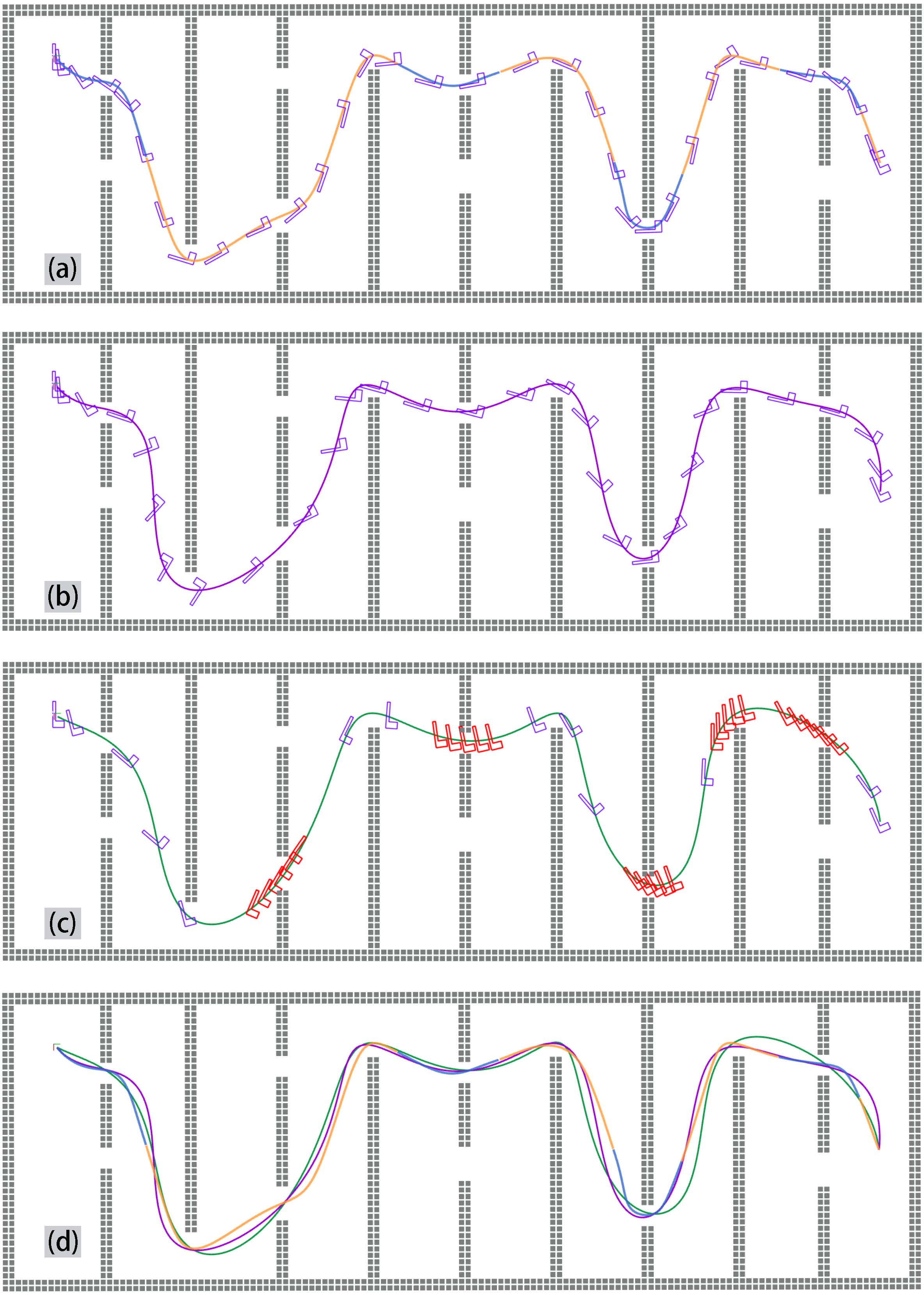}
    \captionsetup{font={footnotesize}}
    \caption{Simulation benchmark of the proposed method against two other any-shape robot trajectory optimization methods in Maze. a) \textbf{Our proposed} method: The SE(2) trajectory pieces (blue curve) and the $\mathbb{R}^2$ trajectory pieces (orange curve) are shown, along with the corresponding L-shaped robot states (purple) at various points. b) \textbf{SVSDF} method (purple curve). c) \textbf{RC-ESDF} method: (green curve), with the red L-shaped robot indicating a collision trajectory segment with obstacles. d) Comparative visualization of the trajectories generated by the three methods.}
    \label{fig:traj_compare}
\end{figure}

\subsubsection{Settings} 
\begin{figure}[htbp]
    \vspace{+2pt}
    \centering
    \includegraphics[width=0.4\textwidth]{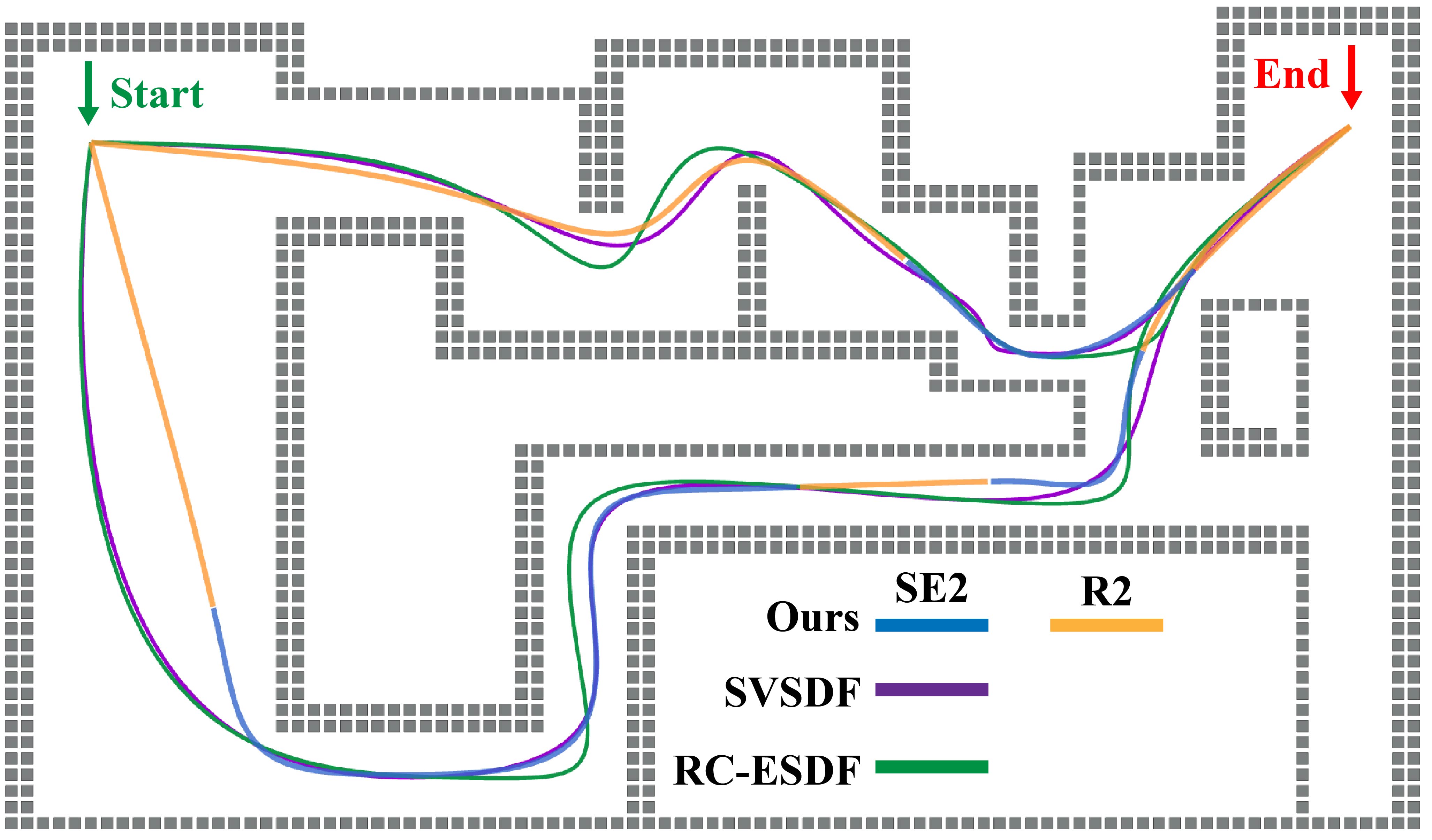}
    \captionsetup{font={footnotesize}}
    \caption{The generated two topological trajectories of the three methods in the \textbf{Office} map.}
    \label{fig:mapA_traj}
    \vspace{-10pt}
\end{figure}

We make the comparison in two customized environments, \textbf{Office} (structured corridors/rooms, as shown in Fig.~\ref{fig:traj_compare}) and \textbf{Maze} (complex labyrinth with multiple homotopy classes, as shown in Fig.~\ref{fig:mapA_traj}). We use two non-convex robot shapes (\textbf{L-Shape} and \textbf{T-Shape}) for each environment. For standardized benchmarking, two representative topological paths per environment are selected: \textbf{Path A} which minimizes physical distance but traverses narrow high-risk regions and \textbf{Path B} which prioritizes safety via longer detours with ample clearance. 

\begin{figure}[htbp]
    \centering
    \includegraphics[width=0.5\textwidth]{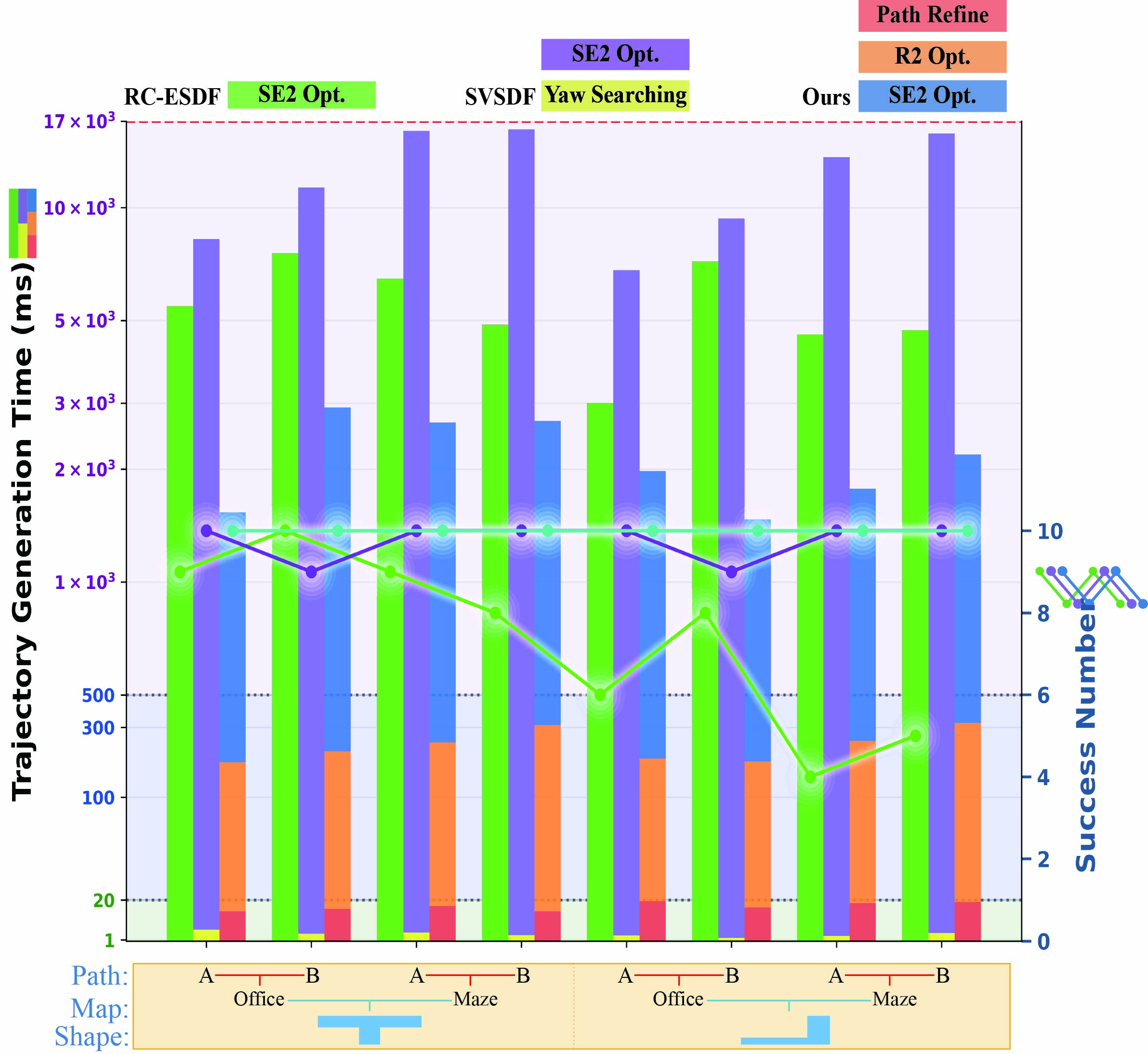}
    \captionsetup{font={footnotesize}}
    \caption{We conducted 10 trajectory generation tests for each robot shape in both maps, showcasing two representative topological paths. The left bar graph illustrates the average time consumed for trajectory generation, while the line graph quantifies the continued collision-free success rates across robots of different shapes.}
    \label{fig:data_graph}
    \vspace{-10pt}
\end{figure}

\subsubsection{Results} As shown in Fig.~\ref{fig:traj_compare} and Fig.~\ref{fig:mapA_traj}, both SVSDF and our method demonstrate effectiveness in CCA while RC-ESDF encounters a collision in the narrow passage of the maze environment (red L-shape in Figure~\ref{fig:traj_compare}).
The computation time and success rate for each method are presented in Fig.~\ref{fig:data_graph}. RC-ESDF directly generates trajectories guided by $\mathbb{R}^2$ paths, achieving shorter computation times than SVSDF but at the cost of lower success rates. This suboptimal CCA performance stems from the ESDF gradient’s failure to consistently indicate optimal collision-avoidance directions during trajectory optimization, particularly for non-convex-shaped robots in maze environments featured with multi-wall. SVSDF improves success rates using an iterative method for implicit computing the swept volume SDF but incurs excessive computational overhead. Our method outperforms both baselines, achieving the highest success rate while maintaining low time costs.

\subsection{Real-world Tests}
We validate our method in a real-world indoor office environment using a robot modified with a goods-shelf structure to emulate non-convex geometry and customer drink delivery capability. The platform employs a Livox Mid360 LiDAR (with built-in IMU) and an Intel NUC onboard computer (i7-1360P CPU). High-accuracy and robust localization and mapping modules (\textit{e.g.}, \cite{fastlivo1,fastlivo2,FASTLIO2}) are necessary for the robot to perform precise obstacle avoidance. We deploy FAST-LIO2 \cite{FASTLIO2} in this work, enabling aggressive maneuvers in narrow spaces. Experimental results (Table~\ref{table:1}) demonstrate the method’s efficiency.

\begin{figure}[!h]
    \centering
    \includegraphics[width=0.45\textwidth]{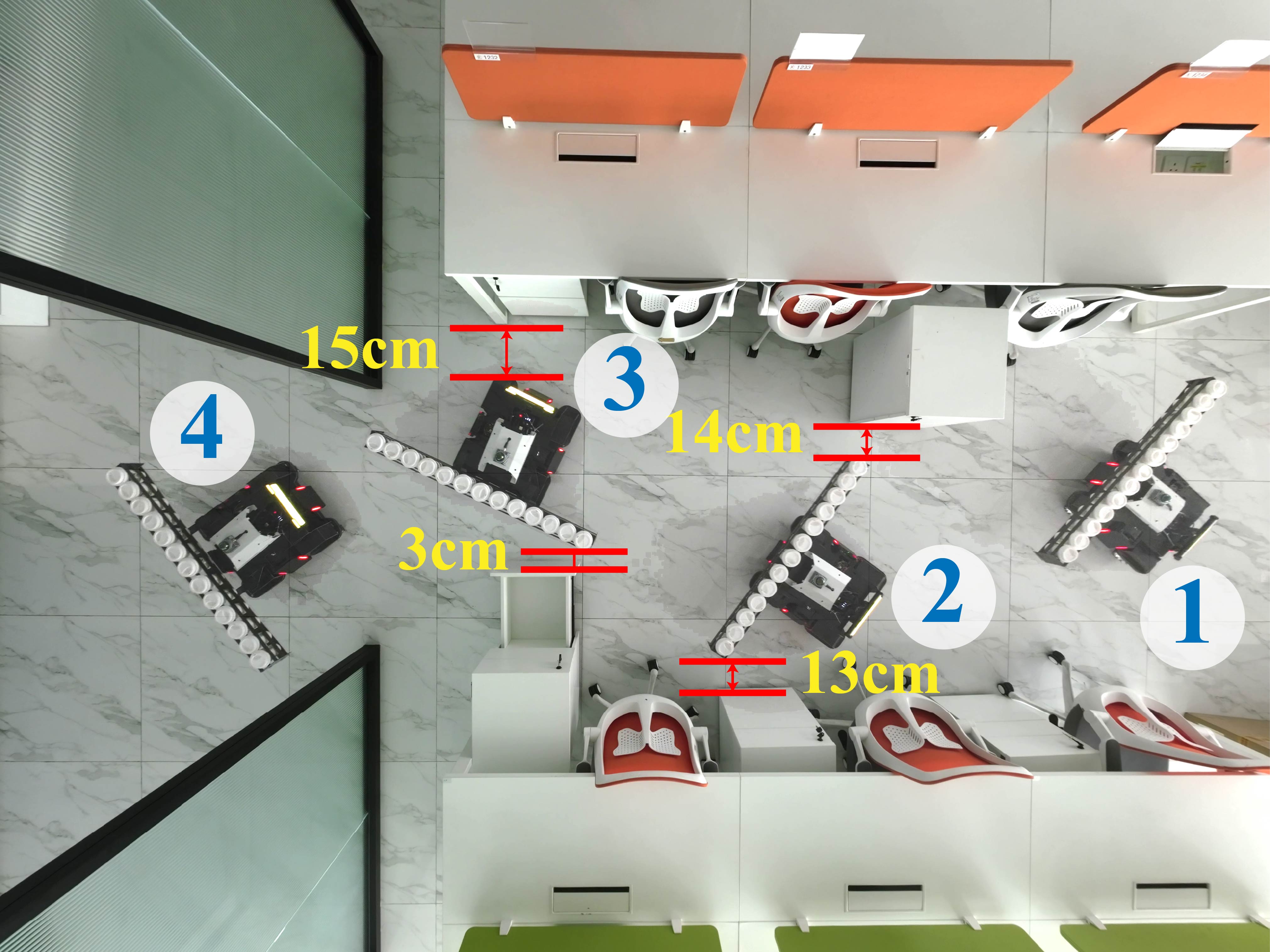}
    \captionsetup{font={footnotesize}}
    \caption{A Bird’s Eye View (BEV) shows a T-shaped robot maneuvering through a narrow, obstacle-filled workspace with irregularly placed cabinets, highlighting the challenge of maintaining safe distances from obstacles across four keyframes.}
    \label{fig:experiment}
    \vspace{-0.3cm}
\end{figure}


\begin{table}[htb]
\centering
\small
\setlength{\tabcolsep}{8pt}
\caption{Real-world Experiment}
\label{table:1}
\begin{tabular}{lccc>{\raggedleft\arraybackslash}c}
\toprule
& Path Refine & \multicolumn{2}{c}{Trajectory} & Total \\
\cmidrule(lr){3-4}
& & $\mathbb{R}^2$ & SE(2) & \\
\midrule
Time (s) & 0.030 & 0.135 & \begin{tabular}[c]{@{}c@{}}11.966\end{tabular} & 12.131 \\
\cmidrule{1-5}
Length (m) & \multicolumn{1}{c}{\begin{tabular}[c]{@{}c@{}}\raisebox{0ex}[0pt]{\scalebox{1.0}{\rotatebox{45}{/}}}\end{tabular}} & \begin{tabular}[c]{@{}c@{}}20.340\end{tabular} & \begin{tabular}[c]{@{}c@{}}34.537\end{tabular} & 54.877 \\
\bottomrule
\end{tabular}
\begin{tablenotes}
    \footnotesize
    \item SE(2) trajectories: Time and length values denote all segments (three pieces total). $\mathbb{R}^2$ trajectories: Optimized as a single piece with total time/length reported.   
\end{tablenotes}
\vspace{-0.5cm}
\end{table}



\section{CONCLUSION AND FUTURE WORK}
In this paper, we presented a novel framework for arbitrary-shaped ground robot planning that possessed both computational efficiency and Continuous Collision
Avoidance feature. Experiments in both simulation and real-world scenarios confirmed the efficiency and practicability of the proposed method. In the future, we will explore the possibility of further deploying this framework in real-time replanning tasks, which could endow the arbitrary-shaped ground robots with the capability of continuous collision-free exploration in unknown environments.

\bibliographystyle{IEEEtran}
\bibliography{citation}

\end{document}